\documentclass{article}
\pdfoutput=1
\usepackage{arxiv}

\usepackage[utf8]{inputenc} 
\usepackage[T1]{fontenc}    
\usepackage{hyperref}       
\usepackage{url}            
\usepackage{booktabs}       
\usepackage{amsfonts}       
\usepackage{amsmath}

\usepackage{nicefrac}       
\usepackage{microtype}      
\usepackage{lipsum}
\usepackage{graphicx}
\usepackage{float}
\usepackage[natbibapa]{apacite}
\graphicspath{ {./images/} }

\usepackage{graphicx}
\usepackage[ruled,vlined,noend]{algorithm2e}
\usepackage{subcaption}
\usepackage{float}

\newcommand{\KL}{D_\mathrm{KL}}

\begin{document}
\title{Understanding the origin of information-seeking exploration in probabilistic objectives for control}
%
\author{
 Beren Millidge \\
  School of Informatics\\
  University of Edinburgh\\
  \texttt{beren@millidge.name}
   \And
  Anil K Seth \\
  Sackler Center for Consciousness Science\\
   Evolutionary and Adaptive Systems Research Group\\
  School of Engineering and Informatics\\
  \And
    Christopher L Buckley \\
  Evolutionary and Adaptive Systems Research Group\\
  School of Engineering and Informatics\\
  University of Sussex\\
  \texttt{C.L.Buckley@sussex.ac.uk} 
  }
%
%
%
\maketitle              

\begin{abstract}

    The exploration-exploitation trade-off is central to the description of adaptive behaviour in fields ranging from machine learning, to biology, to economics. One popular approach to solving this trade-off has been to propose that agents possess an intrinsic 'exploratory drive' whereby agent's are driven to maximize the information they gain about the world through action, as well as the more traditional utility maximization objective. In this paper we mathematically investigate the nature of such objectives and demonstrate that the combination of utility maximizing and information-seeking behaviour arises from the minimization of a class of objectives which we call divergence objectives. We propose a dichotomy in the objective functions underlying adaptive behaviour between \emph{evidence} objectives, which correspond to well-known reward or utility maximizing objectives and our novel \emph{divergence} objectives which instead seek to minimize the divergence between the agent's expected and desired distribution over futures and which give rise to information-seeking exploration. This new class of divergence objectives provides the mathematical foundation for a much richer understanding of the exploratory components of adaptive and intelligent action, beyond simply greedy utility maximization, and can also provide a mathematically elegant explanation for anomalous behaviour often observed in psychophysical and choice tasks such as probability matching.
\end{abstract}

\section{Introduction}

The exploration-exploitation dilemma is a fundamental problem in implementing intelligent, adaptive behaviours in the face of uncertainty \citep{cohen2007should,dayan2008decision,sutton1998introduction,kaelbling1998planning,mobbs2018foraging,berger2014exploration}. When there is substantial uncertainty about the world -- either due to unknown rewards or payoffs of actions, or the intrinsic nonstationarity of the world -- there is always the possibility that the current best policy may in fact be suboptimal, and thus the agent must make a choice between continuing to pursue the current policy (\emph{exploit}), or to try a different policy in the hope that it is better, or else to learn something about the nature of the world (\emph{explore}). However, deliberately selecting currently suboptimal action racks up opportunity cost on average compared to selecting the best action, assuming the action-value estimates are reasonably accurate. Thus, to maximize total reward, it is imperative both to explore sufficiently and efficiently to become certain that the estimates for the returns of actions or policies are correct, and then to exploit the best policy maximally. This problem becomes more acute in environments with changing reward structures where the exploration-exploitation dilemma can never be fully resolved, since the environment, and thus the estimates of value, may always be changing and thus there is always a need to explore \citep{cohen2007should,sun_planning_2011,thrun1991active}. 

The exploration-exploitation trade-off, due to its ubiquity in decision-making under uncertainty, has been studied in a vast number of fields ranging from machine learning and reinforcement learning \citep{thrun1991active,thrun1992efficient,kaelbling1998planning,oudeyer2009intrinsic,sun_planning_2011,schmidhuber1991possibility,schmidhuber1999artificial,schmidhuber2007simple}, cognitive science and neuroscience \citep{cohen2007should,dayan2008decision,geana2016boredom,daw2006cortical,friston_active_2015,gottlieb2013information}, economic decision-making \citep{stigler1961economics,akerlof1978market,macready1998bandit,daw2006cortical,li2008exploration}, organizational theory \citep{march1991exploration,miller2016organizational,gupta2006interplay,sudhir2016exploration,he2004exploration}, evolutionary theory \citep{mehlhorn2015unpacking,vcrepinvsek2013exploration,zhang2019balancing,traulsen2009exploration}, and optimal foraging theory \citep{mobbs2018foraging,kramer1991exploration,charnov2006optimal,monk2018ecology}. Many animals, including humans, often exhibit near optimal choice behaviour in classic bandit tasks which require a solution to the exploration-exploitation dilemma \citep{wilson2014humans,mehlhorn2015unpacking,brenner2006behavior,krebs1978test}, although it is likely that they diverge from optimality on more challenging tasks. In the human brain, the explore-exploit trade-off has been linked to a circuit of brain regions such as the orbitofrontal and anterior cingulate cortices, which appear to compute positive and negative reward valences in respose to events, and the locus coerulus \citep{angela2005uncertainty,yeung2004neural,aston2005integrative}, which appears to be involved in directly mediating a distinction between exploitatory and exploratory behaviour phases through phasic or tonic firing patterns, respectively \citep{sara2009locus,aston1994locus,gilzenrat2003pupil}. Mathematically, an optimal solution is known for the case of n-armed bandits with Gaussian stationary reward distributions, in the form of Gittins indices \citep{gittins1974dynamic}. However, there is no yet fully developed solution to this problem for more complex, and especially nonstationary environments, which characterise many of the environments in which humans and other animals must manage this trade-off \citep{cohen2007should,still_information-theoretic_2012,jamieson2014lil,kolter2009near}.

Recent work in both machine learning and cognitive neuroscience is converging towards understanding the problem of action selection as an inference problem \citep{levine2018reinforcement,friston_deep_2018,friston2003learning,knill2004Bayesian,attias2003planning,toussaint2009probabilistic,kappen2007introduction}. Within neuroscience, this understanding has developed under the aegis of the `Bayesian Brain' framework, which postulates that choice behaviour  can be understood in terms of Bayes-optimal inference. Developments here follow a similar trajectory to perception, which is now often also understood as fundamentally a process of Bayesian inference \citep{knill2004Bayesian,itti2009Bayesian,clark_whatever_2013,sanborn2016Bayesian}, a conceptual heritage which traces back to Helmholtz's ideas of perception as unconscious inference \citep{friston2006free,friston2003learning,friston2012active}. Similarly, in reinforcement learning -- the subfield of machine learning which handles the action selection problem -- formulations recasting standard reinforcement methods as implementing a form of Bayesian inference have become popular and have led to significant algorithmic advances. Under the name of `control as inference' \citep{attias2003planning,toussaint2009probabilistic,toussaint2009robot,rawlik2013probabilistic,theodorou2010generalized,kappen2007introduction,levine2018reinforcement,abdolmaleki2018maximum}, this approach has demonstrated how ubiquitous methods such as Q-learning \citep{watkins1992q}, policy-gradients \citep{williams1992simple}, and actor-critic methods can be recast in such a probabilistic framework and have led to improved algorithms in terms of stability, performance, and ability for simple exploration \citep{abdolmaleki2018maximum,haarnoja2018applications,haarnoja2018soft}. These algorithms also offer improved exploration ability through the use of an `action entropy' term, which encourages the maximization of reward while keeping the distribution over actions as random as possible, which provides a regularising effect. However, the type of exploration such methods provides is fundamentally random, and is thus insufficient in high-dimensional action spaces where purely random exploration is too slow to cover much of the space \citep{tschantz2020reinforcement,shyam2018model,mohamed_variational_2015,houthooft2016vime}. One might expect that these Bayesian approach to action selection, due to their principled Bayesian quantification and treatment of uncertainty, should give rise to powerful \emph{directed} exploration methods which directly leverage the knowledge of uncertainty maintained in the Bayesian framework to directly try to minimize the remaining uncertainty in the problem -- both in terms of the agent's model of the world, its knowledge of its current state, or uncertainty in the mapping between actions and future rewards \citep{tschantz2020reinforcement,ha_recurrent_2018,millidge2020whence,pathak2017curiosity}. However, this has not yet been the case, and probabilistic methods have largely contented themselves with purely random, entropy maximizing exploration \citep{haarnoja2018soft,haarnoja2018applications,rawlik2010approximate,lee2019efficient}. 

However, in both cognitive neuroscience and machine learning, the importance of directed, information-hungry exploration is becoming increasingly recognised. In neuroscience, it is well known that the brain is sensitive to `epistemic affordances' \citep{friston_active_2015,schwartenbeck_computational_2019,friston2017curiosity}. In vision, humans and others are highly sensitive to novelty or \emph{Bayesian surprise} \citep{itti2006Bayesian,itti2009Bayesian,parr2017uncertainty,parr2019computational}, exactly as would be expected from uncertainty-reducing directed exploration and not random exploration. Furthermore, in reinforcement learning, the insufficiency of random exploration, and the necessity for more information-directed exploration in high-dimensional sparse-reward tasks is well understood. The field has experimented with a large number of potential `intrinsic measures' such as information gain and empowerment, and has shown that optimizing these measures can improve performance and lead to more robust and effective artificial agents \citep{haarnoja2018soft,chua_deep_2018,abdolmaleki2018maximum,millidge2019deep,okada_variational_2019}. 

In both fields, however, there is little unified understanding of the origin of these epistemic drives from a principled Bayesian perspective. Typically, empowerment or information gain terms are often derived from a purely intuitive perspective and are optimized either alone or as an ad-hoc addition to a standard objective such as reward maximization. Moreover, there is often no deep understanding of the fundamental objectives that are implicitly being optimized in such a procedure. In the reinforcement learning field, this has led to a profusion of subtly differing algorithms and objectives, without a clear understanding of how they fit together or are related, while in the cognitive neuroscience, it is often unclear what the 'Bayes-optimal' normative objective that biological behaviour should be compared to actually is. In this paper, we aim to provide precisely such a unifying picture. We present a simple unifying framework of Bayesian control objectives by classifying them into two simple types -- Evidence objectives and Divergence objectives. We show that directed, information-maximizing exploration arises naturally from Divergence objectives and \emph{not} from evidence objectives and show that this is due to the entropy-maximizing properties of divergence objectives. Moreover, we show that how both our abstract classes of Evidence and Divergence objectives relate to currently known objectives such as the ELBO \citep{beal2003variational},  control-as-inference \citep{kappen2007introduction,rawlik2013probabilistic}, the expected free energy in active inference \citep{friston_active_2015}, action and perception as divergence minimization \citep{hafner2020action}, and empowerment \citep{klyubin2005empowerment}. 

Crucially, our new framework offers a richer and more detailed framework for understanding the nature of exploratory, information-seeking, behaviour as an intrinsic goal per-se, and the mathematical consequences that entails. Theoretically, we provide the means to go beyond the default assumptions of pure reward, or utility, maximization and instead precisely and mathematically characterise the basis for another type of objective which could be driving behaviour -- \emph{divergence minimization}. Moreover, our theory can parsimoniously explain several puzzling empirical observations from behavioural economics and subjects' behaviour in choice tasks. Specifically, the phenomenon of probability matching \citep{shanks2002re,vulkan2000economist,tversky1974judgment}, where subjects do not always greedily choose the best (highest reward expectation) in stochastic bandit tasks, but instead sample each option in rough proportion to its probability of a reward. For instance, if there is a 90\% option and a 10\% option, utility maximization would predict 100\% selection of the 90\% option, while empirically subjects often sample the 10\% option 10\% percent of the time, which shows that the subject has correctly learnt the outcome probabilities but nevertheless does not implement the utility-maximizing solution. Importantly, minimizing a divergence objective results in precisely this probability matching behaviour. Thus, our mathematical results suggest that this strategy is not pure `irrationality' \citep{tversky1974judgment,vulkan2000economist} but instead arises as a natural result of minimizing a divergence instead of an evidence (utility maximizing) objective and that while this approach performs worse (in terms of pure reward) on simple stationary bandit tasks, in more complex and nonstationary environments with latent structure to learn, the increased exploration from the divergence objective may compensate for these deficits.
\section{Results}

\subsection{Problem Setup}

We consider the control problem at a high level of generality. We assume we have an agent in some sort of environment at a time $t$. We assume no knowledge of the environment except that it has emitted observations $o_{1:t}$ to the agent by time $t$. The agent and environment then run forward to a final time horizon $T$.  The agent can emit actions $a_{1:T}$ which affect future observations $o_{t:T}$. We assume that the agent maintains a model probability distribution over the effect of its future actions on future observations $p(o_{t:T} | a_{t:T})$. The ultimate goal of the agent is to infer some set of actions $a_{t:T}$ that let the agent achieve its goals. Technically, this describes a partially observed Markov-decision-process (POMDP) \citep{kaelbling1998planning,sutton2018reinforcement}

A subtle but important point is how to encode goals, desires, or rewards into the inference procedure. By itself Bayesian inference contains no notion of reward or goals. It simply computes a posterior distribution given a generative model and data. We choose to encode goals or desires into the inference procedure by specifying a separate and exogenous `desire distribution' $\tilde{p}(o_{t:T})$ which is the distribution over observations the agent wishes to obtain \citep{friston2012active}. To obtain an equivalence with common ideas of reward-maximization, it is possible to set $\tilde{p}(o_{t:T}) = \frac{1}{Z}e^{-r(o_{t:T})}$ where $r(o_{t:T})$ is the reward given by a sequence of observations. In effect, this sets the desire distribution to a Boltzmann distribution where the observation granting the largest reward is the most desired, and observations giving less rewards become exponentially less likely in the desire distribution.

We argue that there are two different ways to formally translate the intuitive idea of an agent `achieving its goals' under uncertain future inputs into the mathematical probabilistic framework. Firstly, an agent may attempt to maximize the average \emph{likelihood} of the desire distribution under its expected future observations. We call this the \emph{Evidence Objective}. Secondly, an agent may try to directly match its expected distribution of future observations to its desire distribution, or equivalently minimize the divergence between the two distributions. We call this the \emph{Divergence Objective} \footnote{We maximize and minimize the log of the desire distribution instead of the desire distribution itself due to the better numerical properties of logs. Since $\ln$ is a monotonic function, the log transformation has no effect on the minima of the function. We use $\ln$ to indicate a logarithmic transform. The base of the log is irrelevant.}.
\begin{align}
    \text{Evidence} = \underset{a_{t:T}}{\mathrm{argmax}} \, \, \mathbb{E}_{p(o_{t:T} | a_{t:T})}[\ln \tilde{p}(o_{t:T})] \\
    \text{Divergence} = \underset{a_{t:T}}{\mathrm{argmin}} \, \, \KL[p(o_{t:T} | a_{t:T}) || \tilde{p}(o_{t:T})]
\end{align}

Intuitively, we can think of the evidence objective as simply trying to capture the mode, or the peak of the desire distribution, and concentrating on inferring the actions which concentrate future probability mass around those peaks most effectively. By contrast, divergence objectives encourage precisely matching the expected future distribution and the desire distribution everywhere. If the desire distribution is broad and complex, this means that the expected future distribution under a divergence objective is also encouraged to be broad and complex, while this is not true for an evidence objective, which is encouraged to simply find the largest mode of the desire distribution. This effect can be seen visually in Figure 1. Conversely, in the limiting case of a highly peaked desire distribution, the two objectives become identical.

To gain more mathematical intuition for the differences between the two objectives we can express them in terms of each other.

\begin{align}
    \text{Evidence} &= \underset{a_{t:T}}{\mathrm{argmax}} \, \, \mathbb{E}_{p(o_{t:T} | a_{t:T})}[\ln \tilde{p}(o_{t:T})] \notag \\
    &= \underset{a_{t:T}}{\mathrm{argmax}} \, \, \mathbb{E}_{p(o_{t:T} | a_{t:T})}[\ln \tilde{p}(o_{t:T})\frac{p(o_{t:T} | a_{t:T})}{p(o_{t:T} | a_{t:T})}] \notag \\
    &= \underset{a_{t:T}}{\mathrm{argmax}} \, \, -\underbrace{\KL[p(o_{t:T} | a_{t:T}) || \tilde{p(o_{t:T})}]}_{\text{Divergence}} - \underbrace{\mathbb{H}[p(o_{t:T} | a_{t:T})]}_{\text{Expected Future Entropy}}
\end{align}
By expressing the Evidence objective in terms of the Divergence objective, we can see that it tries to match the two distributions while also \emph{minimizing} the entropy of the expected future distribution. Conversely, we can express the Divergence objective in terms of the Evidence objective.

\begin{align}
    \text{Divergence} &= \underset{a_{t:T}}{\mathrm{argmin}} \, \,  \KL[p(o_{t:T} | a_{t:T}) || \tilde{p}(o_{t:T})] \notag \\ &= \underset{a_{t:T}}{\mathrm{argmin}} \, \underbrace{\mathbb{E}_{p(o_{t:T} | a_{t:T})}[\ln p(o_{t:T} | a_{t:T})]}_{\text{Expected Future Entropy}} - \underbrace{\mathbb{E}_{p(o_{t:T} | a_{t:T})}[\ln \tilde{p}(o_{t:T})]}_{\text{Evidence Objective}}
\end{align}

Intuitively, this decomposition shows that we can think of the divergence objective as trying to optimize the evidence objective whilst \emph{simultaneously} trying to maximize the entropy of the expected future distribution. In effect, a divergence objective tries to keep the future as broad as possible while attaining your goals, while the evidence objective tries to keep the possible futures as narrow as possible while matching your desire distribution. Next, we show that it is this property of maximizing the entropy of your expected future -- or maintaining a broad distribution over possible futures -- that results in information-seeking exploration.

Unlike the evidence objective, the divergence objective immediately decomposes into an information-gain term, which will induce information-seeking directed exploration in an agent that optimizes this term. 

\begin{align}
    \KL[p(o_{t:T} | a_{t:T}) || \tilde{p}(o_{t:T})] &= \mathbb{E}_{p(o_{t:T} | a_{t:T})}[\ln \frac{p(o_{t:T} | a_{t:T})}{\tilde{p}(o_{t:T})}] \notag \\
 &= \mathbb{E}_{p(o_{t:T},x_{t:T} | a_{t:T})}[\ln \frac{p(o_{t:T} | a_{t:T})}{\tilde{p}(o_{t:T})}] \notag \\
 &= \mathbb{E}_{p(o_{t:T},x_{t:T} | a_{t:T})}[\ln \frac{p(o_{t:T},x_{t:T} | a_{t:T})}{\tilde{p}(o_{t:T})p(x_{t:T} | o_{t:T}, a_{t:T})}] \notag \\
 &= \underbrace{\mathbb{E}_{p(x_{t:T})}\KL[p(o_{t:T} | x_{t:T})||\tilde{p}(o_{t:T})]}_{\text{Desire Divergence}} - \underbrace{\mathbb{E}_{p(o_{t:T} | a_{t:T})}\KL[p(x_{t:T} | o_{t:T}, a_{t:T})||p(x_{t:T} | a_{t:T})]}_{\text{Information Gain}}
\end{align}

The reason optimizing the divergence objective results in this elusive information gain functional is due to the fact that the divergence objective maximizes the entropy of future observations given actions. Through a simple information-theoretic identity, we see that maximizing entropy implies the maximization of an information gain between observations and latent variables when a latent variable model is brought into play.
\begin{align}
    \mathbb{H}[p(o_{t:T} | a_{t:T})] &= \mathbb{E}_{p(o_{t:T},x_{t:T} | a_{t:T})}[\ln p(o_{t:T} | a_{t:T})] \notag \\
    &= \mathbb{E}_{p(o_{t:T},x_{t:T} | a_{t:T})}[\ln \frac{p(o_{t:T},x_{t:T} | a_{t:T})}{p(x_{t:T} | o_{t:T}, a_{t:T})}] \notag \\ 
    &=- \underbrace{E_{p(x_{t:T})}\mathbb{H}[p(o_{t:T} | x_{t:T})]}_{\text{Likelihood Entropy}} - \underbrace{\mathbb{E}_{p(o_{t:T} | a_{t:T})}\KL[p(x_{t:T} | o_{t:T}, a_{t:T}) || p(x_{t:T} | a_{t:T})]}_{\text{Expected Information Gain}}
\end{align}

Thus we see that by maximizing marginal observation entropy under a latent variable model, we end up maximizing the information gain between the observations and the latent parameters, while also maximizing the likelihood entropy.

Conversely, since the evidence functional requires \emph{minimizing} the observation entropy, optimizing the evidence functional effectively mandates \emph{minimizing} the information gain between latents and observations. This is the fundamental reason why evidence objectives, and objectives derived from them, do not include information maximizing exploration terms. As we will show below, most objectives optimized in the reinforcement literature derive from variational bounds on evidence objectives, and thus do not demonstrate directed information-maximizing exploration.

\begin{figure}[H]
\centering
\begin{subfigure}{.5\textwidth}
  \centering
  \includegraphics[width=.9\linewidth]{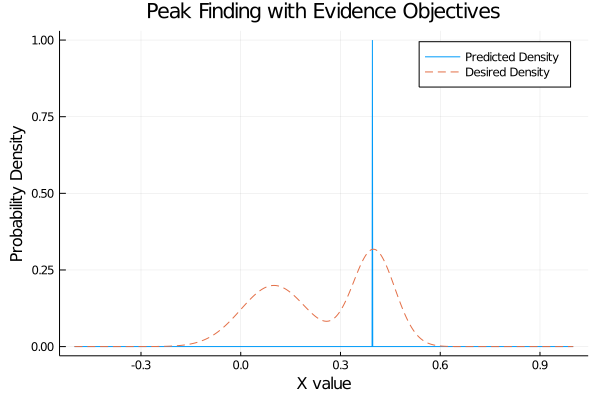}
  \caption{Optimizing with an Evidence Objective}
\end{subfigure}%
\begin{subfigure}{.5\textwidth}
  \centering
  \includegraphics[width=.9\linewidth]{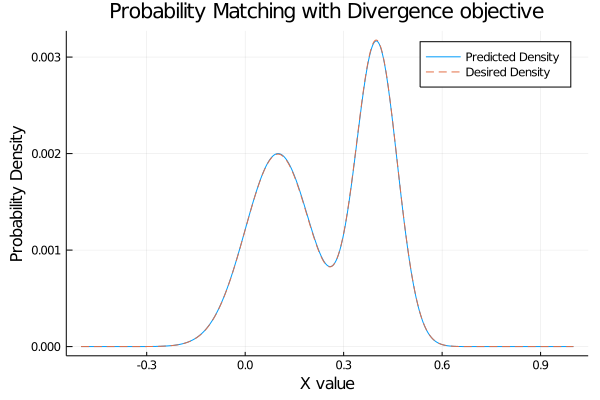}
  \caption{Optimizing with a Divergence Objective}
\end{subfigure}
\caption{Numerical illustration of optimizing a multimodal desired distribution with an Evidence objective (Panel A) vs a Divergence Objective (panel B). The desire distribution consisted of the sum of two univariate Gaussian distributions, with means of $1$ and $4$ and variances of $1$ and $0.4$ respectively. We then optimized an expected future distribution, which also consisted of two Gaussians with free means and variances using both an Evidence and a Divergence objective. As can be seen, optimizing the Evidence Objective results in the agent fitting the predicted future density entirely to an extremely sharp peak around the mode of the desired distribution. Conversely, optimizing a divergence objective leads to a precise match of the predicted and desired distributions (panel B shows the two distributions almost precisely on top of one another). As a technical note, to be able to see both the evidence and desire distributions on the same scale, for the evidence objective the predicted distribution is normalized but the desired distribution is not. Code for these simulations can be found at: https://github.com/BerenMillidge/origins\_information\_seeking\_exploration.}
\end{figure}
\section{Relations to Existing Objectives}
\subsection{Control as Inference}

The control as inference framework \citep{levine2018reinforcement,toussaint2009probabilistic}, although existing for almost two decades in the literature \citep{attias2003planning,theodorou2010generalized,rawlik2013probabilistic,abdolmaleki2018maximum}, has recently come to prominence with the state of the art performance of `soft' actor critic methods \citep{haarnoja2018applications,haarnoja2018soft} in model-free deep reinforcement learning. In effect, these methods regularize standard algorithm such as actor-critic , Q-learning \cite{watkins1992q}, or policy gradients \citep{williams1992simple} with an entropy term which seeks to make the policy as random as possible, thus preventing policy collapse and aiding exploration \citep{levine2018reinforcement}. The control as inference framework conceptualizes control as a problem of Bayesian inference, where the goal is to directly solve the inference problem of computing $p(a_{t:T} | \tilde{o})$. Since this posterior distribution is typically intractable, we approximate it using variational inference.

We define a variational posterior $q(a_{t:T})$ and minimize the KL divergence between the variational distribution and the true posterior. This divergence is also intractable since it contains the true posterior, but a variational bound on this divergence known as the variational free energy (VFE) or evidence lower-bound (ELBO) is tractable, and this bound is optimized \citep{beal2003variational,wainwright2008graphical}.
\begin{align}
    \KL[q(a_{t:T}) || p(a_{t:T} | \tilde{o}_{t:T})] &= \KL[q(a_{t:T}) || \tilde{p}(o_{t:T}, a_{t:T})] - \ln p(o_{t:T}) \notag \\
    &\geq \underbrace{\KL[q(a_{t:T}) || \tilde{p}(o_{t:T}, a_{t:T})]}_{\text{ELBO}}
\end{align}

We can then split up this ELBO into its constituent components.
\begin{align}
    ELBO &= \KL[q(a_{t:T}) || \tilde{p}(o_{t:T}, a_{t:T})] \notag \\
    &= \underbrace{\mathbb{E}_{q(a_{t:T})}[\ln \tilde{p}(o_{t:T} | a_{t:T})]}_{\text{Reward Maximization}} + \underbrace{\KL[q(a_{t:T}) || p(a_{t:T})]}_{\text{Action Complexity}}
\end{align}

We see that the ELBO variational bound splits up into two distinct terms. The first, which contains the desires or goals, functions as a reward maximization term. Broadly, it encourages agents to take actions that will maximize the likelihood of their desires. The second term, action complexity, acts as a regulariser which encourages the variational action distribution $q(a_{t:T})$ to remain as close as possible to the prior action distribution $p(a_{t:T})$. This `action prior' could, for instance, be used to model the intrinsic energetic costs of actions by stipulating that agents should try to use as small actions in absolute size as possible.

In the setting that we take the desire distribution to be a Boltzmann distribution over rewards $\tilde{p}(o_{t:T} | a_{t:T}) = \frac{1}{Z}exp^{-r(o_{t:T}, a_{t:T})}$, and that our action prior $p(a_{t:T})$ uniform, we recover the standard control-as-inference objective as utilized in soft-actor-critic \citep{haarnoja2018applications}, and similar architectures.
\begin{align}
    \mathcal{J}_{CAI} = \mathbb{E}_{q(a_{t:T})}[\ln r(o_{t:T}, a_{t:T}) - \ln q(a_{t:T})]
\end{align}

Next we demonstrate that this term can also be derived as a bound on the evidence objective.
\begin{align}
    \underset{a_{t:T}}{argmax} \ln \tilde{p}(o_{t:T}) &= \underset{a_{t:T}}{argmax} \ln \int dx \, \tilde{p}(o_{t:T} x_{t:T}) \notag \\
    &= \underset{a_{t:T}}{argmax} \ln \int dx \,  \frac{\tilde{p}(o_{t:T} x_{t:T})q(a_{t:T})}{q(a_{t:T})} \notag \\
    &\geq \underset{a_{t:T}}{argmax} \, \mathbb{E}_{q(a_{t:T})}[\ln \frac{\tilde{p}(o_{t:T},a_{t:T})}{q(a_{t:T})}] \notag \\
    &\geq \underset{a_{t:T})}{argmin} \, -\KL[q(a_{t:T}) || \tilde{p}(o_{t:T}, a_{t:T})] \notag \\
    &= \geq \mathcal{J}_{CAI}
\end{align}
We thus see that the control-as-inference objective is derived from an evidence objective and, as expected, does not contain and information-maximizing exploratory terms. The sole force of exploration in this framework arises from the action entropy term which is to be maximized. However, this only gives rise to purely random exploration which is not sufficient in sparse-reward high-dimensional environments.

\subsection{KL Control}

Another method in the reinforcement literature is KL control. Although not widely used, it has received some theoretical analysis \citep{rawlik2013probabilistic,rawlik2013stochastic,lee2019efficient,kappen2005path,kappen2007introduction}. KL control optimizes an objective function,
\begin{align}
    \mathcal{J}_{KL} = \underset{a_{t:T}}{argmin} \KL[p(x_{t:T}) || \tilde{p}(x_{t:T})]
\end{align}
Where we use the notation $x$ instead of $o$ to signify that in the literature KL control has only been applied to systems with directly observable Markovian state as opposed to POMDP dynamics. The KL control objective is clearly a divergence objective and is the only widely used method in the literature to utilize this objective. However, due to its commitment to fully-observed state, it has not been applied to latent variable models and thus its superior exploration capacity has not been well explored in the literature.

\subsection{Active Inference}

Active Inference is a theory of Bayesian action selection in computational and theoretical neuroscience. It extends classical Bayesian brain theories to action and control by viewing action selection as a Bayesian inference problem using a biased generative model \citep{friston2012active,friston2015active,friston2017active,friston2017process,friston2018deep,da2020active,tschantz2020reinforcement,tschantz_scaling_2019,millidge2019deep,millidge2019combining,fountas2020deep}. 

Core to active inference is the \emph{Expected Free Energy} (EFE) functional \citep{friston2015active,parr2019generalised,millidge2020whence,millidge2020relationship} which is optimized through a variational inference procedure in order to select adaptive actions. In active inference, rewards are encoded into a biased generative model of the world $\tilde{p}(o,x)$. This optimization can be formulated as a neurophysiologically realistic messasge-passing scheme for which detailed neuronal process theories have been derived \citep{parr2019neuronal}.

The precise form of the EFE functional underwrites information-seeking directed exploratory behaviour, due to the fact that it can be decomposed into an information-gain `intrinsic value' term and a reward-based 'extrinsic value' term. This decomposition and the complementary decomposition into `risk' and `ambiguity', and its effects on behaviour have been well explored within the active inference literature on a number of standard decision tasks \citep{friston2015active,friston2017process,friston2017curiosity,friston2017graphical}.

The EFE functional can be decomposed in the following two ways,
\begin{align}
    \mathcal{J}_{EFE} &= \mathbb{E}_{q(o,x)}[\ln q(x) - \ln \tilde{p}(o,x)] \notag \\
    &= \underbrace{\mathbb{E}_{q(o,x)}[\tilde{p}(o)]}_{\text{Extrinsic Value}}- \underbrace{\mathbb{E}_{q(o)}\KL[q(x | o) || q(x)]}_{\text{Information Gain}} + \underbrace{\mathbb{E}_{q(o)}\KL[q(x | o) || p(x | o)]}_{\text{Posterior Divergence}} \\
    &= \underbrace{\mathbb{E}_{q(x)}\mathbb{H}[p(o | x)]}_{\text{Ambiguity}} + \underbrace{\KL[q(x) || \tilde{p}(x)]}_{\text{Risk}} + \underbrace{\mathbb{E}_{q(x)}\KL[p(o|x) || \tilde{p}(o|x)]}_{\text{Likelihood Divergence}}
\end{align}

The first decomposition into intrinsic and extrinsic value showcases the information-seeking behaviour engendered by this functional, although the latter decomposition into risk and ambiguity is often the decomposition used in practice in discrete-state-space tasks due to the simpler, more biologically plausible update rules that can be derived \citep{friston2017process,da2020active}. Since the EFE contains the important information-gain term required for directed exploration, we wish to understand how it relates to our framework of evidence and divergence objectives. The relationship of the EFE to the evidence objective (or log-model-evidence) has been extensively studied \citep{millidge2020whence}. We condense and extend their results here while also proving a novel relationship between the EFE and the divergence objective. 

Specifically, we show that the EFE is a lower bound on the evidence objective only when the information gain term is less than the posterior divergence term. Intuitively this occurs when perceptual inference is poor, so that there is a large divergence between true and approximate posterior, but simultaneously much of the environment is known such that there is little information gain remaining in the environment. Over the course of interaction with the environment, it is likely that the optimization of the EFE will converge to exactly the evidence objective, since with an ultimately determinisic environment, and with a sufficiently expressive approximate posterior class, we expect the information gain and the posterior divergence to both converge to 0 in the limit. First, by noting that the extrinsic value of the EFE simply is the evidence objective, we can obtain the relationship:
\begin{align}
    \underbrace{\mathbb{E}_{q(o,x)}[\ln \tilde{p}(o)]}_{\text{Evidence Objective}} &= \underbrace{\mathbb{E}_{q(o,x)}[\ln q(x) - \ln \tilde{p}(o,x)]}_{\text{EFE}} +  \underbrace{\mathbb{E}_{q(o)}\KL[q(x | o) || q(x)]}_{\text{Information Gain}} - \underbrace{\mathbb{E}_{q(o)}\KL[q(x | o) || p(x | o)]}_{\text{Posterior Divergence}} \\
    &\implies \underbrace{\mathbb{E}_{q(o,x)}[\ln \tilde{p}(o)]}_{\text{Evidence Objective}} \geq \underbrace{\mathbb{E}_{q(o,x)}[\ln q(x) - \ln \tilde{p}(o,x)]}_{\text{EFE}} \\
    &\text{If} \, \,  \underbrace{\mathbb{E}_{q(o)}\KL[q(x | o) || q(x)]}_{\text{Information Gain}} \geq \underbrace{\mathbb{E}_{q(o)}\KL[q(x | o) || p(x | o)]}_{\text{Posterior Divergence}}
\end{align}

Since the EFE contains an information gain term to be maximized, it is more likely that it is closely related to a divergence objective. However, here we showcase two derivations that show that the EFE is not directly a bound on the divergence objective either. As with the evidence objective, the EFE only functions as a bound on the divergence objective under certain conditions. The first derivation expresses the bounding relationship between the divergence objective and the EFE in terms of useful interpretable quantities,

\begin{align}
    \underbrace{\KL[p(o) || \tilde{p}(o)]}_{\text{Divergence Objective}} &= \mathbb{E}_{p(o)}[\ln \frac{ \int dx p(o,x)}{\tilde{p}(o)}] \notag \\
    &= \mathbb{E}_{p(o)}[\ln \frac{ \int dx p(o,x)q(x | o)q(o,x)}{\tilde{p}(o)q(x | o)q(o,x)}] \notag \\
    &\geq \mathbb{E}_{p(o)}[\ln \frac{ \int dx p(o,x)q(o,x)}{\tilde{p}(o)q(x | o)q(o,x)}] \notag \\
    &\geq \mathbb{E}_{p(o)}[\ln \frac{ \int dx p(o,x)q(o | x)q(x)}{\tilde{p}(o)q(x | o)q(x | o)q(o)}] \notag \\
    &\geq \underbrace{\mathbb{E}_{p(o)q(x | o)}[\ln q(x) - \ln q(x | o)- \ln \tilde{p}(o)]}_{\text{EFE}}- \underbrace{\mathbb{E}_{p(o)}\KL[q(x | o)||p(o,x)]}_{\text{VFE}} +\underbrace{\mathbb{E}_{q(x | o)p(o)}\KL[q(x | o) || q(x)]}_{\text{Information Gain}}
\end{align}

This shows that the EFE can be expressed as a lower bound on the divergence objective when the VFE $\leq$ the information gain, which is a similar condition to that required for the EFE to serve as a bound on the evidence objective. Similarly, by expressing the EFE in terms of the divergence objective directly, we find that,
\begin{align}
    \underbrace{\KL[p(o) || \tilde{p}(o)]}_{\text{Divergence Objective}} &= \mathbb{E}_{q(x | o)p(o)}\KL[p(o)q(o,x) || \tilde{p}(o)q(o,x)] \notag \\
    &= \mathbb{E}_{q(x | o)p(o)}\KL[p(o)q(o|x)q(x) || \tilde{p}(o)q(x | o)q(o)] \notag \\
    &= \underbrace{\mathbb{E}_{q(x | o)p(o)}[\ln q(x) - \ln \tilde{p}(o) - \ln q(x | o)]}_{\text{EFE}} + \underbrace{\mathbb{E}_{p(o)}\KL[q(x | o) || q(x)]}_{\text{Information Gain}} - \underbrace{\mathbb{H}[p(o)]}_{\text{Marginal Entropy}}
\end{align}

We thus see that the marginal divergence can be expressed as the EFE plus the information gain minus the marginal entropy. This means that the EFE serves as an upper bound (to be minimized) upon the divergence objective only when the information gain is greater than the marginal entropy. This means that while the information about the world which remains to be discovered by the agent is greater than the entropy of an agent's observations, that the EFE faithfully minimizes a bound on the divergence objective, which thus underwrites the exploratory behaviour of active inference agents. However, when the information content in the world falls below the intrinsic entropy of the agent's observations, then the EFE no longer constitutes an upper bound on the divergence objective.
\subsection{Action and Perception as Divergence Minimization}

A recent framework, inspired by active inference and advances in deep reinforcement learning, which aims to unify perception and action under a single framework is Action and Perception as Divergence Minimization \citep{hafner2020action} (APDM). This framework proposes that both action and perception can be modelled as an agent trying to mininimize a divergence functional between two distributions an `actual' distribution $A(x,o)$, and a target distribution $T(x,o)$. 

\begin{align}
    \mathcal{J}_{APDM} &= \KL[A(x,o) || T(x,o)] \\
    &= \underbrace{\mathbb{E}_{A(x)}\KL[A(o | x) || T(o)]}_{\text{Realizing Latent Preferences}} - \underbrace{\mathbb{E}_{A(x,o)}[\ln T(x | o) - \ln A(x)]}_{\text{Information Bound}}
\end{align}

Similar to divergence functionals, the APDM objective can be into a reward-maximization term ('realizing latent preferences') and an `information-bound', which is a bound on the true information gain \citep{agakov2004algorithm} between the actual distribution posterior $A(x |o)$ and the actual distribution prior $A(x)$. 
\begin{align}
    -\underbrace{\mathbb{E}_{A(x,o)}[\ln T(x | o) - \ln A(x)]}_{\text{Information Bound}} &= -\mathbb{E}_{A(x,o)}[\ln T(x | o) - \ln A(x) + \ln A(x | o) - \ln A(x | o)] \\
    &= - \underbrace{\mathbb{E}_{A(o)}\KL[A(x | o) || A(x)]}_{\text{Information Gain}} + \underbrace{\mathbb{E}_{A(o)}\KL[A(x |o) || T(x | o)]}_{\text{Posterior Divergence}}
\end{align}
By expressing this bound explicitly, we can see how it is an upper bound on the information gain, since the posterior divergence is always positive. The tightness of the bound then depends on how closely the actual and target distributions match. In general, we can use this approach to write out a full expression for the divergence objective between two joint distributions over both observations and latent variables.
\begin{align}
    \mathcal{J}_{joint} &= \underset{a}{argmin} \, \KL[p(o,x) || \tilde{p}(o,x)] \\
    &= \underbrace{\mathcal{E}_{p(x)}\KL[p(o| x) || \tilde{p}(o)]}_{\text{Likelihod Divergence}} - \underbrace{\mathbb{E}_{p(o,x)}[\ln \tilde{p}(x | o) - \ln p(x)]}_{\text{Information Bound}} \\
    &= \underbrace{\mathbb{E}_{p(x)}\KL[p(o| x) || \tilde{p}(o)]}_{\text{Likelihod Divergence}} -  \underbrace{\mathbb{E}_{p(o)}\KL[p(x | o) || p(x)]}_{\text{Information Gain}} + \underbrace{\mathbb{E}_{p(o)}\KL[p(x |o) || \tilde{p}(x | o)]}_{\text{Posterior Divergence}}
\end{align}
 
In effect, we see that minimizing the divergence between two joint distributions requires the minimizations of both the likelihood divergence \emph{and} the posterior divergence, while also requiring the maximization of the information between posterior and prior of the first term in the joint KL.

It is also straightforward to relate this joint divergence to the divergence objective, which is the divergence between marginals instead of joints. 
\begin{align}
    \underbrace{\KL[p(o,x) || \tilde{p}(o,x)]}_{\text{Joint Divergence}} &= \KL[p(o)p(x|o)||\tilde{p}(x |o)\tilde{p}(x|o)] \\
    &= \underbrace{\KL[p(o) || \tilde{p}(o)]}_{\text{Divergence Objective}} + \underbrace{\mathbb{E}_{p(o)}\KL[p(x | o)||\tilde{p}(x|o)]}_{\text{Posterior Divergence}} \\
    &\geq \underbrace{\KL[p(o) || \tilde{p}(o)]}_{\text{Divergence Objective}}
\end{align}
Since the posterior divergence is always positive (as a KL divergence), we observe that the joint divergence is simply an upper bound on the divergence objective. Since the divergence is minimized, this bound is in the correct direction, and thus minimizing the joint divergence is a reasonable proxy for minimizing the marginal divergence objective. By minimizing the joint, it implicitly encourages agents to minimize both the marginal divergence as well as the divergence between the predicted and desired posterior distributions.

While the generic APDM divergence, as just a divergence of joints, is straightforwardly an upper bound on the divergence objective, we show that under definitions of the actual and target distributions, the APDM divergence can also be understood as a lower bound on the evidence objective, thus providing a bridge between the two objectives. Although the actual and target distributions can be defined differently depending on the objective you desire to reproduce, one canonical form of the actual and target distributions, which can reproduce control as inference as well as variational perceptual inference is as follows. We define the actual distriution to be the combination of the `real' data distribution $p(o)$ and also a variational belief distribution $q(x | o)$ such that $A(o,x) = q(x | o)p(o)$. Similarly, we define the target distribution to be the product of the agent's veridicial generative model $p(o,x)$ and the exogenous desire distribution $\tilde{p}(o)$ such that $T(o,x) = p(o,x)\tilde{p}(o)$. This target distribution is valid as long as the ultimate objective is optimized via gradients of the divergence, which does not require that the target distribution be normalized. Under this definition of the actual and target distributions, the APDM objective becomes,

\begin{align}
    \mathcal{J}_{APDM} = \KL[q(x | o)p(o) || p(o,x)\tilde{p}(o)]
\end{align}
In the case of known observations in the past, we assume that the data distribution becomes points around the actually observed observations $p(o) = \delta(o = \hat{o})$ while the desire distribution becomes uniform -- as there is little use for control in having desires about the unalterable past. Under these assumptions, the APDM objective simply becomes the ELBO or the negative free energy, thus replicating perceptual inference. However, on inputs in the future, the data distribution becomes a function of action (since actions can change future observations) and the desire distribution becomes relevant, thus allowing the minimization of the APDM functional to underwrite control. To gain a better intuition for the interplay of perception and control in the APDM functional, we showcase the following decomposition,
\begin{align}
    \mathcal{J}_{APDM} &= \KL[q(x | o)p(o) || p(o,x)\tilde{p}(o)] \notag \\
    &= \underbrace{\mathbb{E}_{p(o)}\KL[q(x | o) || p(o,x)]}_{\text{ELBO}} + \underbrace{\KL[p(o)||\tilde{p}(o)]}_{\text{Divergence Objective}}
\end{align}

which demonstrates that the APDM objective effectively unifies action and perception by summing together a perceptual objective (VFE) with the divergence objective for control. This confirms the previous finding that the APDM objective forms an upper bound on the divergence objective since the ELBO, as a KL divergence, is bounded below by 0. We also observe that this form of the APMD objective is also approximately an lower bound on the expected evidence objective, thus providing a link between the two objectives.
\begin{align}
    \mathbb{E}_{p(o | a)}[\ln \tilde{p}(o)] &= \mathbb{E}_{p(o | a)}[\ln \int dx \, \tilde{p}(o,x)] \notag \\
    &= \mathbb{E}_{p(o | a)}[\ln \int dx \, \frac{\tilde{p}(o,x)q(x|o)p(o,x)}{q(x | o)p(o,x)}] \notag \\
    &\geq \mathbb{E}_{p(o)q(x | o)}[\ln \frac{\tilde{p}(o,x)p(o,x)}{q(x|o)p(o,x)}] \notag \\
    &\geq \mathbb{E}_{p(o)q(x | o)}[\ln \frac{\tilde{p}(o)\tilde{p}(x|o)p(o,x)}{q(x|o)p(o)p(x|o)}] \notag \\
    &\geq -\underbrace{\mathbb{E}_{p(o)q(x | o)}[\KL[q(x|o)p(o)||p(o,x)\tilde{p}(o)]}_{\text{APDM Objective}} +\underbrace{\mathbb{E}_{p(o)q(x|o)}[\ln \tilde{p}(x|o) - \ln p(x|o)]}_{\text{Posterior Divergence Bound}} \notag \\
    &\approx \geq \underbrace{\mathbb{E}_{p(o)q(x | o)}[\KL[q(x|o)p(o)||p(o,x)\tilde{p}(o)]}_{\text{APDM Objective}}
\end{align}

Which is approximately equal to the APDM objective under the condition that the posterior divergence bound between desire posterior $\tilde{p}(x | o)$ and true posterior $p(x | o)$ is small.

\subsection{Empowerment}

A number of works have studied the empowerment function in relation to exploration and reinforcement learning \citep{oudeyer2009intrinsic,klyubin2005empowerment,baumli2020relative,jung2011empowerment} as well as in biological systems. Formally, empowerment can be defined as
\begin{align}
    \mathcal{J}_{empowerment} = \underset{a_{t:T}}{argmax} \, \, \mathcal{I}[x_{t:T}, a_{t:T} | a_{1:t}, x_{1:t}]
\end{align}

which intuitively encourages agents to maximize the amount of information about the future contained in the present and past. We can think of this as encouraging agents to `keep their options open' as well as obtain control over the future dynamics, to render them predictable given current and past state. We show that an empowerment objective arises naturally as one component of the divergence functional, when extended to maintain a choice distribution over action, and also show that this decomposition reveals the complement of empowerment -- which we call filtering information gain, which tries to maximize information about the past given the future. Intuitively, this additional term encourages the agent to explore the future so as to better understand the past. To simplify notation, we split up timesteps into past and future, so that we denote $p(o_{1:t}) = p(o_<)$ and $p(o_{t:T}) = p(o_>)$.

We then write out the divergence objective.
\begin{align}
    \KL[p(o_{1:T})||\tilde{p}(o_{1:T})] &= \KL[p(o_{1:T},x_{1:T},a_{1:T})||p(x_{1:T},a_{1:T} | o_{1:T})\tilde{p}(o_{1:T})]  \notag \\
    &= \KL[p(o_> | x_>)p(x_>, a_> | x_<, a_<)p(x_<, a_< | o_<)p(o_<) || p(x_>, a_> | o_>)p(x_< | o_<, x_>, a_>)\tilde{p}(o_>)\tilde{p}(o_<)] \notag \\
    &= \underbrace{\mathbb{E}_{p(x_> | a_>)}\KL[p(o_> | x_>) || \tilde{p}(o_>)]}_{\text{Future Divergence}} - \underbrace{\mathbb{E}_{p(x_<, a_<,o_<)}\KL[p(x_>, a_> | o_>)|| p(x_>, a_> | x_<, a_<)]}_{\text{Generalized Empowerment}} \notag \\ &- \underbrace{\mathbb{E}_{p(o_>, x_<, a_<)}\KL[p(x_<, a_< | o_<, x_>, a_>)||p(x_<, a_< | o_<)}_{\text{Latent Filtering Information}} + \underbrace{\KL[p(o_<) || \tilde{p}(o_<)]}_{\text{Past Divergence}}
\end{align}

At any given time $t$, the prior distribution over observations is fixed $p(o_<) = \delta(o_< = \hat{o}_<$) where $\hat{o}_<$ is the sequence of realized observation values actually observed. When this is substituted into the past divergence term, it becomes just an entropy of the past desire distribution. If we assume that the past desire distribution is constant, then the past divergence term vanishes entirely.

The remaining terms are the `future divergence' term which encourages the agent to match the distribution of future observations expected given the current state with the future desire distribution. In reinforcement learning this term would represent reward maximization in the future. The two remaining terms which are \emph{maximized} are the `generalized empowerment' which is the mutual information between the future actions and states expected given observations and the `prior' expectation of future actions and states. In effect this term underwrites active exploration in state, action, and model-parameter space. It encourages the agent to seek out future observations which will render future actions and states maximally predictable, over and above what could be predicted from previous states and actions. This will also lead to a drive towards empowerment by taking the actions which lead the future environments and states to be maximally controllable but different from the expected future states and actions given past states and actions.

The final remaining term is the `latent filtering information' term which is the complement to the generalized empowerment term from above. Like the empowerment term, this is an information gain term which is to be maximized. In effect, this term encourages agents to maximize the information between the estimates of past states given knowledge of future states and without such knowledge. In effect, this term encourages agents to select the actions that will place it in future states which are informative about past states. Mathematically, this corresponds to the operation of filtering -- or deriving new information about the past from the future. An intuitive situation where this term applies is if there is a rat in a maze which pulls a lever which can either unlock door A or door B. This term would then drive the rat to explore which door is unlocked, so as to gain new information about the effect of the previous lever pulling.

Intuitively, we can think of this term as encouraging the agents to discover the future consequences of its past actions. When phrased in this way, it becomes apparent that the effect of this term may be to encourage the agent to build explicitly \emph{causal} models of action, so that the effects of past actions can be estimated, and then action plans that can verify or disprove these estimates can be selected.


\section{Discussion}

In this paper we have taken a step towards providing a mathematically principled foundation for understanding the origin of information-gain or information-seeking exploratory terms in control problems. Specifically, we have investigated the origin of information gain terms arising from mathematically principled foundations. We have discovered two classes of potential objectives for control, which we call \emph{Evidence} and \emph{Divergence} objectives; where Evidence objectives try to maximize the expected desire distribution to the distribution of future observations, while Divergence objectives instead try to match the future observartion distribution to the desire distribution. We have shown that information gain terms arise naturally out of divergence objectives but \emph{not} evidence objectives, demonstrating that to obtain information-seeking exploration in a control objective, a divergence objective is required. This is due to the fact that divergence objectives mandate the maximization of the marginal entropy of future observations, which encourages agent's to keep their distribution of future observations as broad as possible while still maximizing the likelihood of the desire distribution. Through a simple information-theoretic identity, we have shown that maximizing this entropy, when the entropy is extended to include latent variables, requires maximizing the information gain about the latent variables. When latent variables include things such as unobserved states or model-parameters, this information gain term induces information-seeking exploratory behaviour as a natural consequence of minimizing a divergence objective.

In the literature there is a large number of different `intrinsic measures' or exploration-inducing objectives which can be optimized in control tasks to encourage exploration and ultimately improve performance by better addressing the exploration-exploitation dilemma \citep{storck1995reinforcement,schmidhuber1991possibility,schmidhuber2007simple,oudeyer2009intrinsic,chentanez_intrinsically_2005}. Such objectives include prediction-error maximization \citep{pathak2017curiosity},policy entropy \citep{rawlik2010approximate,haarnoja2018soft}, state entropy \citep{lee2019efficient}, variational information gain \citep{houthooft2016vime,okada_variational_2019,kim2018emi,shyam_model-based_2019}, variational bounds on empowerment \citep{gregor2016variational,karl2017unsupervised}, divergence between ensembles \citep{chua_deep_2018,shyam_model-based_2019} and uncertain state bonuses \citep{bellemare2016unifying,o2017uncertainty}. Cataloging and relating all these different objectives is a mammoth task, which we have only just started in this paper. Nevertheless, understanding how all these functionals relate to one another and to an overarching abstract framework such as the one presented in this paper, is crucial to gain a full understanding of the landscape of possible functionals, and the effects of each on behaviour \citep{oudeyer2009intrinsic}. Moreover, such an understanding can encourage new and potentially more effective objectives to be developed, or can suggest mathematically principled enhancements to existing functionals which can empirically improve their performance \citep{millidge2020whence,noel2021online}. A clear example of this in the literature is recasting of standard model-free RL as probabilistic variational inference under the control-as-inference framework, which has lead to significantly improved model-free RL algorithms \citep{rawlik2010approximate,haarnoja2018applications,levine2018reinforcement}.


Beyond reinforcement learning, this abstract framework for control problems has deep implications for results in cognitive psychology, psychophysics, and choice behaviour. Our work framework suggests that such information-seeking behaviour emerges from a fundamental objective to minimize the divergence between distributions, rather than to simply maximize the evidence or utility. Such intrinsically motivated information-seeking behaviour has been observed and discussed in a wide range of fields, and especially cognitive science. It is well-known, for instance, that saccades in visual foraging are sensitive to regions of the visual scene expected to be informative, as well as task relevant regions \citep{friston_active_2017, itti2006Bayesian,itti2009Bayesian,yang2016theoretical}. This behaviour is easily explained and unified as the optimization of a divergence functional. Similarly, any task which requires the combination of information-seeking exploration and goal-directed action can be straightforwardly modelled as being the result of a divergence-minimizing objective.  This approach replaces the previously ad-hoc practice of simply augmenting a reward-maximizing or goal directed objective with additional 'intrinsic value' terms by putting such a process on a principled mathematical footing. 

Finally, our approach may have deep implications for understanding empirically observed behaviour in behavioural choice tasks in cognitive science and economics. For instance, the much-derided `probability matching' strategy humans often exhibit in choice tasks \citep{daw2006cortical,tversky1974judgment,gaissmaier2008smart,shanks2002re,wozny2010probability,west2003probability} where instead of always choosing the stochastic reward with the highest mean, participants tend to select options in proportion to their probability, can be straightforwardly explained through the minimization of a divergence, instead of an evidence, objective. Interestingly, the intuition that probability matching is not just pure `irrationality', but instead corresponds to a strategy intrinsically sensitive to epistemic goals has been raised before in the literature \citep{vulkan2000economist,shanks2002re}, often conceptualized as an inductive bias towards finding patterns \citep{wozny2010probability,gaissmaier2008smart}, or else as a direct informational bias towards learning in nonstationary environments \citep{burns2009chance}. Our framework provides a precise mathematical grounding of such intuitions; namely that in more complex tasks with a hidden latent variable structure, divergence objective minimization naturally induces intelligent, information-seeking behaviour, which will benefit the agent more in the long run due to the additional exploration performed than with a purely greedy objective. Using the divergence functional becomes slightly sub-optimal from a utility-maximization standpoint in tasks which do not require exploration, since in effect the utility maximizing evidence objective is being regularised by a marginal divergence term, which results in probability matching. However, the advantage of minimizing a divergence objective is that in changing environments, where exploration, and especially information-seeking exploration is vital, the divergence functional will outperform the greedily utility-maximizing approach. While we have focused primarily on the mathematical theory in this article, by providing such a well-characterised framework which makes a number of behavioural predictions, we hope that more empirical research is done to establish whether humans or other agents actually do tend to optimize strictly with evidence maximizing, or divergence maximizing functionals, or indeed whether their behaviour cannot be fully characterised by either side of our dichotomy.


\bibliography{cites}
\end{document}